\documentclass[10pt,twocolumn,letterpaper]{article}

\usepackage{cvpr}
\usepackage{times}
\usepackage{epsfig}
\usepackage{graphicx}
\usepackage{amsmath}
\usepackage{amssymb}

\usepackage[breaklinks=true,bookmarks=false]{hyperref}

\cvprfinalcopy 

\def\cvprPaperID{****} 

\begin{document}

\title{TernausNetV2: Fully Convolutional Network for Instance Segmentation}

\author{Vladimir I. Iglovikov\\
Level5 Engineering Center, Lyft Inc\\
Palo Alto, CA 94304, USA\\
{\tt\small iglovikov@gmail.com}
\and
Selim S. Seferbekov\\
Veeva Systems\\
Frankfurt am Main, 60314, Germany\\
{\tt\small selim.seferbekov@veeva.com}
\and
Alexander V. Buslaev\\
Mapbox R$\&$D Center\\
Minsk, 220030, Belarus\\
{\tt\small aleksandr.buslaev@mapbox.com}
\and
Alexey A. Shvets\\
Massachusetts Institute of Technology\\
Cambridge, MA 02139, USA\\
{\tt\small shvets@mit.edu}
}

\maketitle

\begin{abstract}
The most common approaches to instance segmentation are complex and use two-stage networks with object proposals, conditional random-fields, template matching or recurrent neural networks. In this work we present TernausNetV2 - a fully convolutional network that allows extracting objects from a high-resolution satellite imagery on an instance level. The network has popular encoder-decoder type of architecture with skip connections but has a few essential modifications that allows using for semantic as well as for instance segmentation tasks. This approach is universal and allows to extend any network that has been successfully applied for semantic segmentation to perform instance segmentation task. In addition, we generalize network encoder that was pre-trained for RGB images to use additional input channels. It makes possible to use transfer learning from visual to a wider spectral range. For DeepGlobe-CVPR 2018 building detection sub-challenge, based on public leaderboard score, our approach shows superior performance in comparison to other methods. The source code and corresponding pre-trained weights are publicly available at \url{https://github.com/ternaus/TernausNetV2} 
\end{abstract}

\section{Introduction}
Automatic building extraction from high-resolution satellite imagery creates new opportunities for urban planning and world population monitoring. Traditionally, the building boundaries are delineated through manual labeling from digital images in the stereo view using the photogrammetric stereo plotters \cite{san2010building}. However, this process is a tedious task and requires qualified people and expensive equipment. For this reason, building extraction using the automatic techniques has a great potential and importance. The advantages of satellite imagery compared to aerial imagery are the almost worldwide availability and that the data typically contains wider spectral range, that includes both optical, infrared and extra channels. The geometric resolution of 0.3-1.0 m per pixel is worse than for aerial imagery, but is sufficient to be able to extract large objects, such as buildings. The worldwide availability of the data makes it possible to produce topographic databases for nearly any region of the earth. 

In the last years, different methods have been proposed to tackle the problem by creating convolutional neural networks (CNN) that can produce a segmentation map for an entire input image in a single forward pass. One of the most successful state-of-the-art deep learning method is based on the Fully Convolutional Networks (FCN) \cite{long2015fully}. The main idea of this approach is to use CNN as a powerful feature extractor that creates high-level feature maps. Those maps are further upsampled to produce dense pixel-wise output. The method allows training CNN in the end to end manner for semantic segmentation with input images of arbitrary sizes. This method has been further improved with skipped connections and now known as U-Net neural network \cite{ronneberger2015u}. Skip connections allow combining low-level feature maps with higher-level ones, which enables precise pixel-level localization. A large number of feature channels in upsampling part allows propagating context information to higher resolution layers. This type of network architecture proved itself well in a satellite image analysis competitions. \cite{goldberg2018urban, iglovikov2017satellite, zhang2017building}. Another modification to the U-Net architecture that lead to a first place in the Carvana Image Masking Challenge \cite{kaggle_carvana} was to replace encoder by a first few convolution blocks of the VGG11 network. This modification was called TernausNet \cite{iglovikov2018ternausnet} that we naturally extend in the current work (see also \cite{shvets2018automatic, shvets2018angiodysplasia}).

The semantic segmentation is not able to separate different instances because the predicted boundaries are usually not fine and closely packed objects of the same class collapse into one connected component. It may also happen that there is no distance between objects at all and even perfect network will predict different instances as being part of the same connected blob. In work, \cite{bai2017deep} authors propose a method that utilizes three stacked networks, the first one performs semantic segmentation, the second one predicts gradients of the distance transform, the last predicts energy levels that are used in the postprocessing step during the watershed transformation. Our method is similar in the spirit, but much more straightforward. 

In this work, we solve two different problems. First of all, we  use all available multispectral information. Then, we need a way to modify the network, so that combination of its outputs  allows to make segmentation on the instance level. To resolve the first problem, we suggest an extension of the TernausNet architecture \cite{iglovikov2018ternausnet} that replaces VGG11 encoder with a more powerful ABN WideResnet-38 \cite{bulo2017place}. We also extend the input RGB channels to 11 multispectral channels. So that, we are able to perform transfer learning from RGB to RGB + multispectral inputs. For the second issue, we use ideas that were developed in winning solutions for recent data science challenges \cite{goldberg2018urban, dsbowl2018}. To be specific and separate buildings in a predicted binary masks, we add additional output channel that predicts areas where objects are touched or close to each other. This output is used in a post-processing step and allows to partition the mask into separate instances. 

\section{Dataset}
\begin{figure}[t]
\begin{center}
\includegraphics[width=\linewidth]{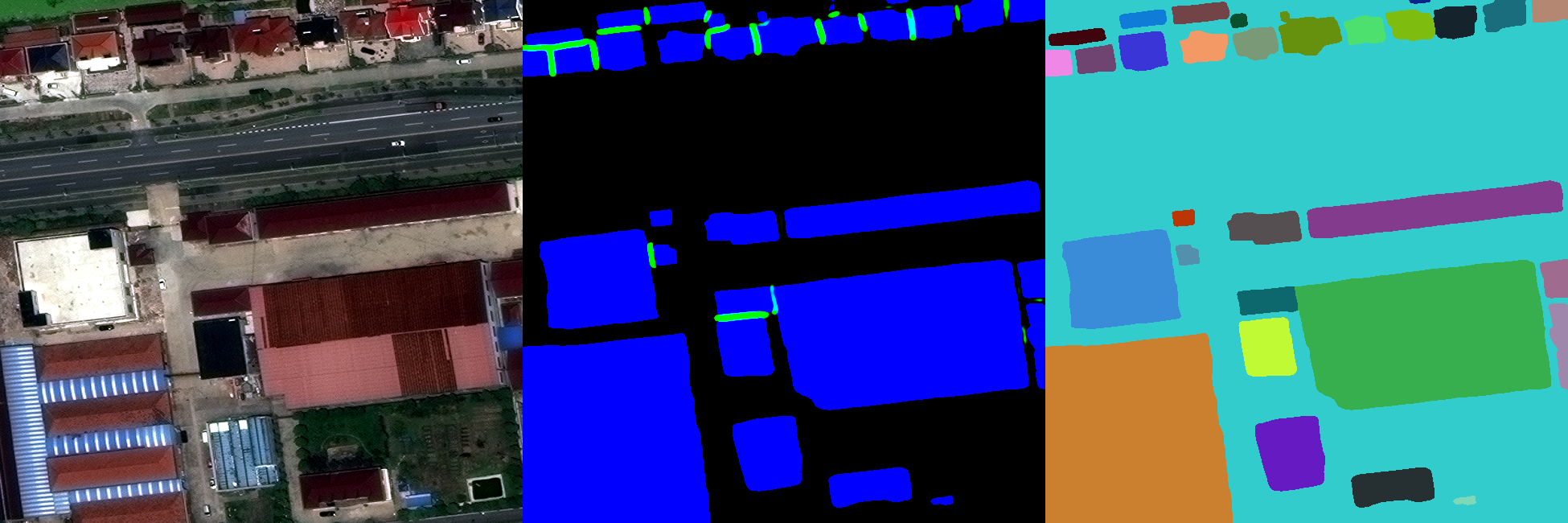}
\end{center}
   \caption{From left to right: RGB part of the input image, predicted binary mask in blue and touching borders  in green, building instances after the watershed transform.}
\label{fig:buildings}
\end{figure}

The training data for the building detection sub-challenge originate from the SpaceNet dataset \cite{spacenet_dataset}. The dataset uses satellite imagery with 30 cm resolution collected from DigitalGlobe's WorldView-3 satellite. Each image has 650x650 pixels size and covers 195x195 $m^2$ of the earth surface. Moreover, each region consists of high-resolution RGB, panchromatic, and 8-channel low-resolution multi-spectral images. The satellite data comes from 4 different cities: Vegas, Paris, Shanghai, and Khartoum with different coverage, of  (3831, 1148, 4582, 1012) images in the train and (1282, 381, 1528, 336) images in the test sets correspondingly. All images in the train set have a paired list of polygons that describes building instances. The labels are not perfect due to the cost of mask annotation, especially in places with high density. To evaluate our model performance the predicted masks for the test images should be upload into DeepGlobe website \cite{deepglobe_website, demir2018deepglobe}. An example of a test image and predictions of our method is depicted in the Fig. \ref{fig:buildings}. \ref{fig:buildings}.

\begin{figure*}
\begin{center}
\includegraphics[width=\textwidth,height=9cm]{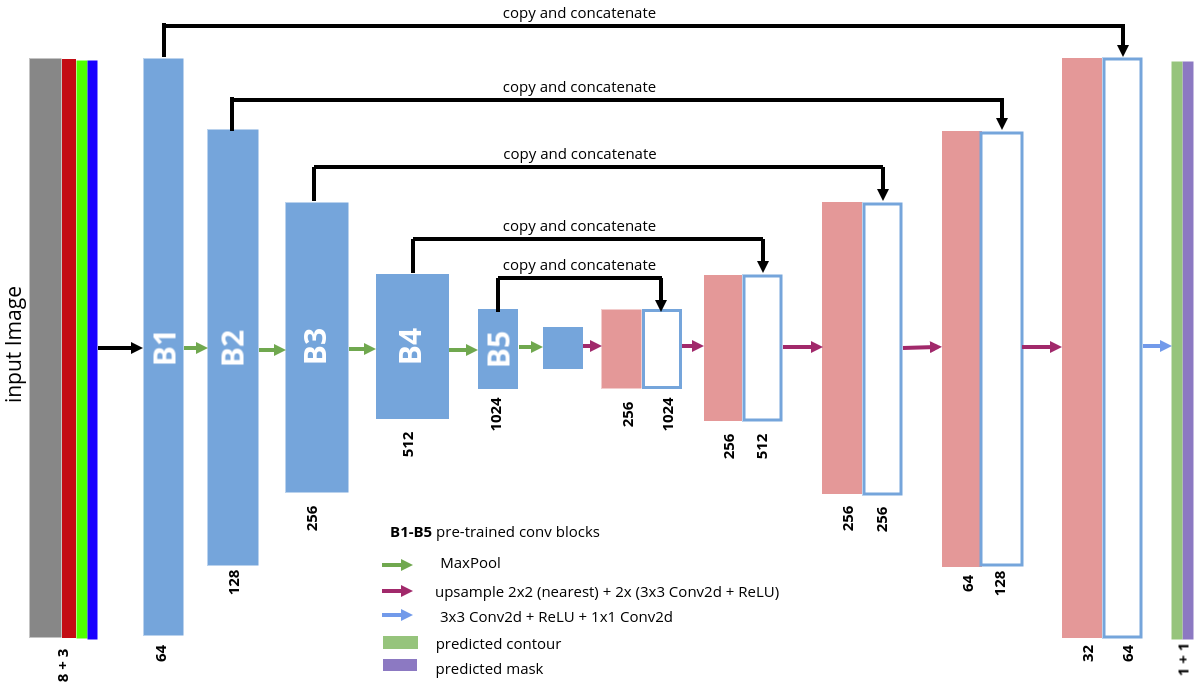}
\end{center}
 \caption{TernausNetV2: encoder-decoder network with skipped connections that has ABN WideResnet-38 as the encoder. As an input, we have RGB + extra channels image. B1-B5 are the first five convolutional blocks of the base network that was pre-trained on the ImageNet. At every step of the decoder block, we perform upsampling, followed by the series of the convolution layers. Skip connections are added between convolution blocks in the encoder and the decoder of the corresponding size. In the end, 1x1 convolution is added to reduce the number of channels to the desired two, one for the binary mask and another one for touching instances.
   }
\label{fig::fpn}
\end{figure*}

\section{Model}
Our approach leverages encoder-decoder type architecture with skipped connections that is also known as U-Net \cite{ronneberger2015u}. In general, U-Net consists of a contracting path to capture context and of symmetrically expanding path. This enables precise localization with skip connections added between blocks of the same size in the contracting and expansive parts. Skip connections allow information to flow directly from the low level to high-level feature maps without alternations that even further improve localization accuracy and speed up convergence  \cite{ronneberger2015u}. The contracting path follows the typical architecture of a convolutional network with alternating convolution and pooling operations and progressively down samples feature maps, increasing the number of feature maps per layer at the same time. Every step in the expansive path consists of an upsampling of the feature map followed by a series of convolution layers. The output of the model is a pixel-by-pixel mask that outputs the class of each pixel. 

As an improvement over the U-Net architecture, we replace the encoder with a convolutional part of the WideResNet-38 network with in-place activated batch normalization \cite{bulo2017place} that was pre-trained on ImageNet. In-place activated batch normalization merges batch normalization layer with activation layers which lead to up to 50\% memory savings. It allows to fit into the GPU memory larger batches and work with the input images of the larger size.  A model based on this encoder showed state of the art performance on semantic segmentation tasks for Mapillary Vistas \cite{neuhold2017mapillary} and Cityscapes \cite{cordts2016cityscapes} datasets. Compared to the original ResNet architecture \cite{he2016deep}, WideResnet uses layers with a higher number of channels, while reducing the number of layers. We use the first five convolutional blocks of the network as an encoder. The decoder of our network consists of five decoder blocks that are connected to the corresponding encoder block of the same size. The transmitted block from the encoder is concatenated to the corresponding decoder block. Each decoder block contains two sets of 3x3 convolutions, followed by ReLU activations \cite{glorot2011deep} that is followed by an upsampling layer that increases the size of the feature map twice. To prevent artifacts at the edges of the predicted buildings, we use nearest neighbor upsampling that showed the best result in our experiments. The output of the model is a two-channel pixel-by-pixel image where the first channel contains a binary mask of the combined building footprint. The second channel contains building borders that are attached to each other or separated by few pixels (see Fig. \ref{fig:buildings}).

To allow the encoder that was pre-trained on RGB images to take 11 channels as an input (RGB + 8 multispectral), we replace the first convolutional layer by a larger one. So that, it takes 11 channel images as an input. We copy weights of the original pre-trained WideResnet38 to the first three channels and 
initialize the remaining channels by zeros.

\section{Training}
The satellite imagery in the Spacenet dataset comes in an 11-bit format. In order to make pixel intensity distributions closer to the usual RGB images, we perform min-max normalization per channel $(x - x_{min}) / (x_{max} - x_{min})$. Then, we normalize RGB part subtracting (0.485, 0.456, 0.406, 0, 0, 0, 0 , 0, 0, 0, 0) and dividing by (0.229, 0.224, 0.225, 1, 1, 1, 1, 1, 1, 1, 1) for each channel correspondingly.

During a training, to perform a smooth transition from RGB to RGB + multi-spectral data we train our network with the following schedule. At the first epoch, we freeze all weights in the encoder, so that only weights in the decoder are trained. Because weights that correspond to the extra layers are zero-initialized only RGB part of the input is used during training. At the end of the first epoch, decoder weights have meaningful with respect to the problem values. At the second epoch, we unfreeze all layers and train it end to end. As a result, the network learns how to go from three to a larger number of input channels in a delicate, careful manner, slowly increasing weights of the multi-spectral part of the input.

As an output of the network, we have an image with two channels. These channels are independent, and both of them need to predict binary masks. One for building footprints and the second one for touching borders. As a loss function, we use a combination of a binary cross entropy and a soft Jaccard loss. This loss was inspired by \cite{iglovikov2017satellite} where authors proposed a way to generalize discrete Jaccard index (also known as intersection over union) into a differentiable form. This allows the network to optimize the loss directly during the training process. 

The Jaccard index can be interpreted as a similarity measure between a finite number of sets. For two sets $A$ and $B$, it can be defined as following:
\begin{equation}
\label{jaccard_iou}
    J(A, B) = \frac{|A\cap B|}{|A\cup B|} = \frac{|A\cap B|}{|A|+|B|-|A\cap B|}
\end{equation}
Since an image consists of pixels, the last expression can be adapted for non-discrete objects in the following way:
\begin{equation}
\label{dicrjacc}
J=\frac{1}{n}\sum_{c=1}^2w_c\sum\limits_{i=1}^n\left(\frac{y_i^c\hat{y}^c_i}{y_{i}^c+\hat{y}^c_i-y_i^c\hat{y}_i^c}\right)
\end{equation}
where $y_i^c$ and $\hat{y}_i^c$ are a binary values (label) and corresponding predicted probability for the pixel $i$ of the class $c$. For simplicity, we choose $w_1 = w_2 = 1$. 

An image segmentation task can also be considered as a pixel classification problem. We additionally use common classification loss function for a binary cross entropy, denoted as $H$ that we apply independently to each output channel.

The final expression for the generalized loss function is obtained combining Eg. (\ref{dicrjacc}) and $H$ as following:
\begin{equation}
\label{free_en}
L=\alpha H +(1-\alpha)(1-J)
\end{equation}
By minimizing this loss function, we simultaneously maximize predicted probabilities for the right class for each pixel and maximize the intersection over union $J$ between masks and corresponding predictions. In our experiments we used $\alpha = 0.7$. 

As an additional regularization, we apply extensive data augmentation both spatial and in the color space. For spatial augmentation we use a random re-size, randomly choosing scale between 0.5 and 1.5 of the input image and mask. We apply random rotations in the full (0, 360) range, using reflection padding if needed. From the resulting image and mask, we crop random regions of the size 384x384 pixels. These images are subject to color transformations such as random contrast/brightness and gamma corrections with gamma coefficient randomly chosen between two discrete values: 0.8 and 1.2. One video card GTX1080 Ti with 11 GB of memory allows using the batch size of 5 images. In our case, we use 4 GTX1080 Ti and batch 20. 

We train our network using Adam optimizer with learning rate 1e-4. The training is done for 800 epochs. At the inference time, we make predictions on the whole image padding it with 11 pixels on each side to the 672x672 size, so that it would be divisible by $32=2^5$ (5 is the number of max-pooling layers in the decoder that constrains the allowed input sizes). After prediction is done the padded regions is cropped. 

The last step during the inference is to post process predicted binary masks and touching borders in such a way that the binary mask is splitted into separate instances. To make this, we subtract touching borders from the corresponding mask to obtain seeds and use both masks and these newly generated seeds as an input to the watershed transform. We do not fine tune the model for different cities. We also do no use bagging, checkpoint averaging, test time augmentations or any other ensembling techniques in our solution. The end to end process including network inference and watershed transformation process ten samples per a second using one GTX 1080 Ti.

\section{Conclusions}
We developed a model for satellite imagery building detection. We used a fully convolutional neural network that is traditionally used for semantic segmentation and added additional output that adds instance segmentation functionality. As an encoder, we chose pre-trained on ImageNet WideResnet-38 network with in-place activated batch normalization that can generate good semantic features and it is memory efficient at the same time. We also generalized this pre-trained encoder and propose training schedule that allows applying transfer learning from RGB to multi-spectral data. Based on the public leaderboard score our model provides state of the art result with the score equal to 0.74.

\section*{Acknowledgment}
The authors would like to thank Open Data Science community \cite{ods_website} for many valuable discussions and educational help in the growing field of machine/deep learning.

{\small
\bibliographystyle{ieee}
\bibliography{landlib}
}

\end{document}